# Evaluating Discussion Boards on BlackBoard© as a Collaborative Learning Tool:

A Students' Survey and Reflections[1].

Abdel-Hameed A. Badawy
Electrical and Computer Engineering Department,
University of Maryland,
College Park, MD USA
absalam@umd.edu

Michelle M. Hugue
Computer Science Department,
University of Maryland,
College Park, MD USA
meesh@umd.edu

*Abstract* In this paper, we investigate how the students think of their experience in a junior level course that has a blackboard course presence where the students use the discussion boards extensively. A survey is set up through blackboard as a voluntary quiz and the student who participated were given a freebie point. The results and the participation were very interesting in terms of the feedback we got via open comments from the students as well as the statistics we gathered from the answers to the questions. The students have shown understanding and willingness to participate in pedagogy-enhancing endeavors.

*Keywords-collaborative learning; evaluation; discussion boards; Blackboard; pedagogy*

---

[1]A link to the IEEE published paper can be found here: 10.1109/ICEMT.2010.5657540

## I. INTRODUCTION

The students in both computer science have to take a C language programming [1] course that is worth four credits which involves a lot of programming. This course requires a lot of effort from the teaching staff to help the students with their debugging issues and to clarify ambiguous, unclear or even unintentional mistakes in the requirements of the to-be delivered projects. This c programming course has a website where lecture slides, project descriptions and other related documents can be retrieved. One of the courses that follow the c programming course is a regular three credits course that teaches computer organization, which also has a significant programming component in it as well. In some cases, projects can be as large as several hundreds of lines of code.

The follow-up course is taught using the support of course management software. The course management software adopted by the University of Maryland at College Park (UMD) is called Enterprise Learning Management System (ELMS) [3] and is powered by Blackboard version 8.0 at the time of conducting this study. Blackboard version 9.1 was just released recently [2].

TABLE I. DISTRIBUTION OF ACCESSES PER EACH DISCUSSION FORUM.

| Forum | Accesses | % Total | Students | Staff | Ratio |
|---|---|---|---|---|---|
| Lab 1 | 4154 | 30% | 2756 | 1398 | 66% |
| Lab 2 | 3332 | 24% | 3241 | 91 | 97% |
| Lab 3 | 4557 | 33% | 3985 | 572 | 87% |
| Lab 4 | 876 | 6% | 876 | 0 | 100% |
| Technical | 833 | 7% | 700 | 133 | 84% |
| Total | 13752 | 100% | 11558 | 2194 | 84% |

TABLE II. DISTRIBUTION OF MESSAGES PER EACH DISCUSSION FORUM.

| Forum | Messages | % of Total | Students | Staff | Ratio |
|---|---|---|---|---|---|
| Lab 1 | 69 | 25% | 44 | 25 | 64% |
| Lab 2 | 73 | 27% | 69 | 4 | 95% |
| Lab 3 | 69 | 25% | 61 | 8 | 88% |
| Lab 4 | 32 | 12% | 32 | 0 | 100% |
| Technical | 31 | 11% | 24 | 7 | 77% |
| Total | 274 | 100% | 230 | 44 | 84% |

There exists a discussion board that students can use for instructor created topics. In case of the course under consideration here, each programming project had its own discussion forum. There were four projects. In addition, there was a forum for technical questions related to accessing or using the computer resources for the course. The traffic on the discussion boards for the first project was overwhelming. Therefore, we decided to design a survey to be able to get the feedback of the students about their experience with the discussion boards, its effectiveness and its contribution to their learning. An earlier compressed version of the survey was presented in an internal UMD teaching and learning conference [4].

## II. MOTIVATIONAL STATISTICS

To motivate the purpose of our study, we have used the existing statistics collection tools in ELMS which collects usage statistics for all parts of the courseware to show how many posts happened during the semester and how many times the posts were accessed and/or read.

Tables (I) and (II) show the usage statistics of the forums in terms of total number of accesses to each forum and the number of unique posts and/or responses in each forum. In addition, we have broken up each of these according to the total number of events belonging to the students as opposed to the total number of forums belonging to the instructional staff.

Each table shows six columns. The first column of both tables (I) and (II) shows the name of the different forums. There were a total of five forums in this course. There exists one forum per each programming laboratory (a total of four programming labs) and a technical questions forum where they would ask/inquire about any computer related questions to each other or to the instructional staff. The second column in table (I) and (II) shows the number of accesses/messages per each forum respectively. Column three shows the percentage of the accesses/messages of each forum with respect to the total number of accesses/messages to all the forums. Columns four and five show the distribution of each forum accesses/messages with respect to who view/wrote them whether it is the students or the instructional staff. Column six shows the percentage of the students accesses/messages to the overall number of accesses/messages for each forum.

Examining the number reported in tables (I) and (II), one cannot ignore the very large number of accesses to the forums, which are almost 14K accesses relative to the very small number of messages exchanged on the forum of less than 300. Using very rough/simple math, the ratio of the number of accesses per posted message is 50 to 1 i.e. per each message posted there is an average of 50 views to it. To get even more interesting, we got also statistics about the number of accesses and number of message. To keep things in perspective, one have to keep in mind that the total number of students enrolled in this class was 65 students. There was four instructional staff for this course.

We can compute the following statistics:
1) On average, the number of messages posted per student is about 3.
2) On average, the number of messages read per student is 178.
3) On average, the number of messages posted per staff member is 11.
4) On average, the number of messages read per staff member is 549.
5) The ratio of the number of messages posted by all the students to the messages posted by all the staff members is about 5.
6) The ratio of the number of messages read by students to the messages read by all the staff members is about 5.

One can conclude several conclusions from the averages and ratios that we have computed so far. First, the student read far more messages than what they write or post. This means that with careful monitoring for the on-going flow of messages/posts on the forums we can reach the students with crucial clarification answers to hairy questions that we know that many of the students will read.

Another very important issue to notice here is that the staff members read three times more messages compared to the students, which makes sense. The instructional staff is working hard to follow the discussion to make sure the correct information is disseminated among the students by their fellow students.

The statistics above suggest that the forums on ELMS are really a collaborative learning tool for the students. The students were the real source of most of the traffic on the forums. The students asked and answered their own questions except in the rare cases where one of the instructional staff had to step in and correct or rectify a problematic issue. Clearly, the forums were a running archive for the students showing what the questions that were asked previously are and they were able to ask further questions. We can conclude here that the forums are a form of self-online office hours that are run by the students and monitored by the instructional staff.

### III. SURVEY PARTICIPATION AND LOGISTICS

The statistics we have shown thus far show how much the forums help the students as per the total number of accesses and all the other statistics that we computed so far. In order to increase the participation and reward the students who will participate in the survey, we gave each participating student a freebie point to be added to his/her total earned points. This amounts to a one point on a scale of 100.

The survey was finished by 47 students out of the total of 65 class students and was attempted by 53 students. We consider this great response from the students since the goal at the UMD for the course evaluations done at the end of the semester of 70% participation. We had a 72% completion rate and an 82% attempting rate. The survey was open for participation from the students for a period of 36 hours only. It is worthy of noting here that the students of our class have scored a large participation turnout of over 90% in the campus-wide course evaluation which shows that we could have gotten better participation rate if the students were given more time to turn their surveys in.

### IV. SURVEY RESULTS

Tables (III) through table (XI) summarize the results that we have obtained from our survey. The first cell on the top left columns of each table is the question the table is addressing. Table (XII) shows that we had asked the students for their comments for all questions except for question seven and eight. The amount of comments that we got were in many cases very thoughtful statements. We will go over some of the very interesting ones. We will compile all of the comments into an appendix and will publish it as a technical report. Please contact the first author if you are interested in obtaining the full set of student comments.

TABLE III. SURVEY QUESTION(1).

| Do you think that the ELMS website is a helpful learning tool in this course? | % |
|---|---|
| Yes, it is a helpful learning tool. I love it. | 64% |
| No, it is not a useful learning tool. I hate it. | 2% |
| I neither love it nor hate. I am neutral. | 34% |

TABLE IV. SURVEY QUESTION(2).

| Do you prefer courses that use ELMS over other courses with a regular website? | % |
|---|---|
| Yes, courses are better with ELMS. | 63% |
| No, I like non-ELMS courses better. | 6% |
| It does not really matter. I do not care | 31% |

TABLE V. SURVEY QUESTION(3).

| How often do you post on the discussion boards? | % |
|---|---|
| Once Daily. | 2% |
| Once a week. | 19% |
| Once a month. | 21% |
| Once a semester. | 26% |
| Never posted. | 13% |
| I only read but I do not post. | 45% |

TABLE VI. SURVEY QUESTION(4).

| If you have questions, do you prefer to go to office hours or you try the boards first? | % |
|---|---|
| I prefer ELMS board posts. | 36% |
| I prefer to talk to someone face to face. | 43% |
| Depends on the time I have to figure out the answer. | 36% |
| I just ask a classmate. | 21% |

TABLE VII. SURVEY QUESTION(5).

| If you posted to the board and a fellow student answered your question, do you trust his answer? | % |
|---|---|
| Yes, sure I trust my classmates. | 53% |
| No, they might be wrong. | 9% |
| Only if someone from the instructional staff says it is a fine. | 26% |
| Not on all issues I trust my classmates' answers. | 21% |

TABLE VIII. SURVEY QUESTION(6).

| How many posts do you read? | % |
|---|---|
| I just read everything. | 40% |
| It is a waste of time. I read nothing. | 2% |
| I read posts depending on the title of the post. | 51% |
| I read posts related to my questions only. | 17% |

TABLE IX. SURVEY QUESTION(7).

| If there is a course with two sections one without an ELMS website and another with an ELMS website, which section would you enroll in? | % |
|---|---|
| The ELMS-based section. | 43% |
| The non-ELMS section. | 6% |
| It is not a factor at all. | 51% |

TABLE X. SURVEY QUESTION(8).

| Currently you can't submit anonymous questions and /or answers to the boards. If there was that option, would you participate more by reading and/or writing? | % |
|---|---|
| I would have participated more. | 30% |
| It wouldn't matter to me. | 55% |
| I wouldn't trust the posts if they were anonymous. | 15% |

TABLE XI. SURVEY QUESTION(9).

| Did you participate in this survey because of the freebie point? | % |
|---|---|
| Yes | 89% |
| No | 11% |

TABLE XII. NUMBER OF COMMENTS PER SURVEY QUESTION.

| Question # | # of comments |
|---|---|
| 1 | 40 |
| 2 | 35 |
| 3 | 28 |
| 4 | 25 |
| 5 | 28 |
| 6 | 25 |
| 7 | N/A |
| 8 | N/A |
| 9 | 31 |

V. A SELECTION OF THE STUDENTS' COMMENTS

In this section, we will share some of the student comments that are very thoughtful and interesting. Here are a selected few of the comments of the students.

- "I do like giving feedback on pedagogy. Student feedback is hard to get/give in a big lecture hall. It benefits and improves the quality of the educational environment".
- "ELMS is alright, but if all the features were used, it could be (even) better".
- "I would have taken the survey without the point".
- "I also do feel that this survey is important".
- "Since there is no anonymity (in the boards), I am fairly certain no one would knowingly pass on false information (since their name associated with the post)".
- "If a teacher puts lecture notes and other materials on ELMS is helpful. Other teachers of mine has said they would use ELMS and then after 2 weeks they just gave up and just used emails".
- "Discussion boards are the most useful tool".
- "In the boards, I ask general questions to draw from the knowledge of the entire class rather than focusing all questions to TA's and the professor".
- "I think it would be helpful if all courses use ELMS".
- "(ELMS is) also nice for viewing grades and the syllabus and assignments".
- "Someone can post answers that are not perfect but still he/she knows better than I do".
- "If I can see the reasoning (in a classmate's answer of a question) and it makes sense, I can trust their answer well enough".
- "It (ELMS and its associated Discussion boards) makes classes without discussion sections easier to follow".
- "I do wish they would upgrade (the ELMS) GUI, it feels like a 90's web app".
- "Some classes do not use ELMS very well and others are very organized and utilize it".
- "(My issues with ELMS): you can't save your password to log into elms through google chrome which is painful".
- "(ELMS is) a great idea, but the interface is terrible".
- "When I have a question that is not answered by the discussion boards, I assume it is a question only I have and I will then go to a teacher to ask it".
- "Almost always, the face-2-face benefits are better than ELMS".
- "It's usually easier and more convient to post a message to the discussion board".

VI. RELATED WORK

The volume of work that is being pursued in the last few years in education and the theory of teaching and learning is beyond any single person to follow up but nevertheless there is a lot of work that is related to our work here. For example, Leung and Ivy discuss the usefulness of course websites and they present and articulate the perception of the students of these course websites [7]. A study of students' perception in New Zealand considered different online learning strategies to determine which ones are more valuable [8]. Gokhale examined collaborative learning techniques in a study and has shown that collaboration among students enhances learning and increases critical thinking [9]. Coopman has examined the effectiveness and best practices of using Blackboard as an e-learning environment [10].

## VII. CONCLUSIONS

It is very clear to us that the current pattern of use of ELMS had made our students think of ELMS as equivalent to a class webpage and an associated discussion boards. We as instructors should use other features of ELMS like the blog, voice discussion boards, and/or voice email. Our students think that the interface of ELMS is too old looking and it needs a facelift. The students do not want the use of ELMS in addition to other web presence for the course since it is distracting.

Our students trust us (their instructors) as well as their classmates for the most part. We should trust our students and give them the chance to give us their view of the world as often as possible. The students gave us more feedback than what we hoped for and imagined. The students do want to help make the education process better for people who come after them. Our students are mostly millennial students and we should have that in mind and expect initiative, technology sophistication and willingness to participate especially if they see their instructors do care about them [5,6].

## ACKNOWLEDGMENT

The authors would like to acknowledge all the staff at the UMD's Center for Teaching Excellence (CTE) especially Henrike Lehnguth, Dave Eubanks and Prof. Spencer Benson. Without their feedback and support, neither this paper nor the Innovations in Teaching and Learning conference presentation would have materialized. In addition, we would like to thank Ms. Mary Maxson of the R. H. Smith School of Business' Information Technology for sharing a technology use survey done at the School of Business at UMD.